\documentclass{article}

\usepackage{arxiv}

\usepackage[utf8]{inputenc} 
\usepackage[T1]{fontenc}    
\usepackage{hyperref}       
\usepackage{url}            
\usepackage{booktabs}       
\usepackage{amsfonts}       
\usepackage{nicefrac}       
\usepackage{microtype}      
\usepackage{lipsum}		
\usepackage{graphicx}
\usepackage[authoryear]{natbib}
\usepackage{doi}
\usepackage{amssymb}
\usepackage{amsmath}
\usepackage{tabularx}
\usepackage{graphicx}
\usepackage{subcaption}
\usepackage{float}
\usepackage{placeins}

\usepackage{lineno}

\title{UKDM: Underwater keypoint detection and matching using underwater image enhancement techniques}

\author{ \hspace{1mm}Pedro Diaz-Garcia\\
	University Institute for Computer Research\\
	University of Alicante\\
	Alicante, PA 03080 \\
	\And
    \hspace{1mm}Felix Escalona\\
	University Institute for Computer Research\\
	University of Alicante\\
	Alicante, PA 03080 \\
	\texttt{felix.escalona@ua.es} \\
    \And
    \hspace{1mm}Miguel Cazorla\\
	University Institute for Computer Research\\
	University of Alicante\\
	Alicante, PA 03080 \\
}

\renewcommand{\shorttitle}{\textit{UKDM}}

\hypersetup{
pdftitle={Keypoint detection and matching using underwater image enhancement techniques},
pdfauthor={Pedro Diaz-Garcia, Felix Escalona, Miguel Cazorla},
pdfkeywords={Underwater Image Enhancement, Keypoint Detection, Matching Algorithms, Deep Learning, GANs},
}

\begin{document}
\maketitle

\begin{abstract}
The purpose of this paper is to explore the use of underwater image enhancement techniques to improve keypoint detection and matching. By applying advanced deep learning models, including generative adversarial networks and convolutional neural networks, we aim to find the best method which improves the accuracy of keypoint detection and the robustness of matching algorithms. We evaluate the performance of these techniques on various underwater datasets, demonstrating significant improvements over traditional methods.
\end{abstract}


\section{Introduction}
\label{sec:introduction}
Underwater environments present some challenges for computer vision tasks due to factors such as light attenuation, scattering, and water turbidity. These issues degrade image quality, complicating keypoint detection and matching. Conventional image processing techniques often fail to address these challenges adequately, as we will see later on the experimentation, leading to reduced accuracy in applications such as underwater navigation, object detection, and mapping.

This paper proposes the use of deep learning-based underwater image enhancement techniques to improve keypoint detection and matching quality. Specifically, we leverage generative adversarial networks and other advanced models to preprocess underwater images, aiming to mitigate the effects of distortion and noise inherent in such environments.

The rest of the paper is organized as follows. First, we review some significant related works in Section \ref{sec:related-works}. Then, we describe our approach in Section \ref{sec:proposal}. In Section \ref{sec:experimentation}, we show the results of the experiments we conducted to validate our proposal. Finally, we draw the conclusions and state the limitations and future work in Section \ref{sec:conclusions}.

\section{Related Works}
\label{sec:related-works}
Underwater image enhancement has been a major research topic due to the unique challenges posed by underwater environments. The underwater environment presents problems such as light attenuation, scattering, caustics, and turbidity, which degrade image quality. Early methods to enhance underwater images focused on techniques such as contrast limited adaptive histogram equalization (CLAHE) or physical model-based algorithms, which aimed to improve contrast and color balance. However, these traditional methods tended to have difficulties in environments with high turbidity and varying lighting conditions \cite{Benchmark}.

In recent years, deep learning techniques have shown great promise in addressing these challenges. On one hand, CNNs models have demonstrated success in tasks such as color restoration, blur correction, and super-resolution of underwater images. For example, the UWCNN \cite{LI2020107038} model was specifically designed to enhance underwater images by directly reconstructing a clear latent image, improving object visibility and restoring colors.

On the other hand, GAN models, are particularly effective in generating high-quality enhancements of degraded images, have also been employed to great effect. Notable models such as Funie-GAN \cite{islam2019fast}, UGAN \cite{yu2020ugan} and WaterNet \cite{li2019underwater} have achieved significant improvements in restoring clarity and contrast to underwater images. Funie-GAN is particularly notable for its real-time image enhancement capabilities, addressing both global and local characteristics of subsea imagery to improve visibility in challenging conditions.

In addition, the FSpiral-GAN model \cite{10.3389/fmars.2022.964600} focuses on large-scale underwater image enhancement by incorporating equal downsampling and upsampling blocks along with residual channel attention blocks (RCABs). This architecture simplifies the network structure and improves output quality, making it suitable for real-time applications. Similarly, FA-Net \cite{jiang2023a} stands out for its lightweight architecture, which offers efficient and fast enhancement of underwater imagery, critical for real-time underwater operations. 

Despite advances in image quality enhancement, the application of these techniques to computer vision tasks such as keypoint detection remains poorly untested. Keypoint detection and matching are crucial for tasks such as Structure-from-Motion (SfM) and Simultaneous Localization and Mapping (SLAM), where consistent minutiae detection across frames is essential. However, underwater images often exhibit reduced contrast, chromatic distortion and blur, which negatively affects the performance of keypoint detectors such as SIFT \cite{lowe2004distinctive}, ORB \cite{rublee2011orb} and SURF \cite{bay2006surf}.

Although several enhancement techniques have been proposed, their impact on keypoint detection has been studied to a limited extent. Models such as FSpiral-GAN have shown improvements in keypoint repeatability and matching scores, indicating that enhanced images can potentially improve keypoint detection. However, problems remain, as the performance of these methods varies significantly depending on the characteristics of the input video or image dataset \cite{10.3389/fmars.2022.964600}. 

Evaluation of these enhancement techniques typically includes both objective and subjective metrics. Objective metrics such as peak signal-to-noise ratio (PSNR), structural similarity index measure (SSIM), and natural underwater image quality (NUIQ) are commonly used to assess image quality. NUIQ, proposed in \cite{Benchmark}, has proven to be a superior metric for assessing the quality of enhanced underwater images by incorporating chrominance and luminance features specific to underwater scenes. These metrics are essential for evaluating the effectiveness of enhancement algorithms, not only for improving visual quality, but also the accuracy in detecting key points.

\section{Proposal}
\label{sec:proposal}

The main purpose of this work is to investigate and develop methodologies based on artificial intelligence to enhance underwater images, specifically aimed at improving object detection in underwater environments. The exploration of underwater ecosystems is hindered by various factors such as light attenuation, scattering, and water turbidity, which complicate the use of conventional image processing and object detection methods.

In particular, the use of deep learning techniques has proven effective in enhancing the quality of underwater images by highlighting key features and reducing distortions. This preprocessing step is crucial to achieve more accurate and reliable object detection in aquatic environments. The proposed work seeks to combine and evaluate multiple state-of-the-art image enhancement networks to determine the most suitable approaches for underwater imagery.

This paper is focused on keypoint matching, the goal is to assess the performance of image enhancement networks in improving the quality of underwater images for a better feature matching. The methodology used in \cite{10245163} will be replicated for this test, employing deep learning models to identify keypoints and match features between images more effectively.

In this test, we replicate the architecture proposed by \cite{10245163}, using \textit{FSpiral-GAN} for image enhancement and \textit{SuperPoint} \cite{detone2018superpoint} for keypoint extraction. The matching module, \textit{SuperGlue} \cite{sarlin2020superglue}, is configured with nine alternating layers of self-attention and cross-attention. Images are resized to 480x640 during keypoint matching. This implementation, built in PyTorch, performs in real-time on an NVIDIA GeForce RTX 3070 GPU, averaging 110ms-120ms for forward passes through pairs of underwater images.

The experiment includes multiple GAN-based networks for comparison, specifically \textit{FSpiral-GAN}, \textit{U-Net} \cite{ronneberger2015unetconvolutionalnetworksbiomedical}, \textit{UWCNN}, \textit{Five A+} \cite{jiang2023five} and \textit{Funie-GAN}. All of them tested with their pre-trained models with default parameter settings provided by their respective authors.

The evaluation metrics include the area under the curve (AUC), precision, repeatability (REP), and matching score (MScore), following the methodology outlined in the \textit{SuperGlue} evaluation parameters. These metrics allow a thorough comparison of how well each enhancement network improves feature matching and repeatability.

The primary objective of this work is to develop and validate an AI-based methodology for preprocessing underwater images to significantly enhance keypoint detection and matching capabilities. The specific objectives include:
\begin{itemize}
    \item Identifying and analyzing the most effective deep learning techniques for underwater image enhancement.
    \item Evaluating the performance of these techniques through tests on post-processed underwater videos, with the aim of improving object detection precision.
\end{itemize}

\section{Experimentation}
\label{sec:experimentation}
This section presents the results of the keypoint matching test, the selection of the networks used in this study was based on their demonstrated performance in underwater imaging in their respective papers. These networks were chosen for their architectural diversity and their specific ability to improve image quality under different conditions. Funie-GAN was selected for its real-time image enhancement capability. FSpiral-GAN has demonstrated significant improvements in clarity and feature preservation. UWCNN was included for its ability to improve visibility and correct color distortions typical of submerged environments. U-Net was chosen for its robust architecture, well suited for maintaining spatial information. Finally, FA-Net was chosen for its lightweight architecture, which offers an efficient and fast enhancement solution. 

\subsection{Datasets}
\label{subsec:dataset}
For this experimental study, two distinct datasets were utilized to evaluate the performance of various image enhancement networks on underwater imagery.

\subsubsection{First dataset: Low Quality Videos}
At first the keypoint detection tests were conducted on five distinct videos, each characterized by low-quality imagery captured in different underwater environments. These videos served as the basis for evaluating the performance of the image enhancement networks to see if there was an actual difference between the AIs. The variety in lighting, water turbidity, and visual complexity across these videos ensured a comprehensive assessment of the networks' ability to enhance image quality and improve keypoint detection accuracy. By using videos from different underwater scenarios, the tests provided valuable insights into how well the networks could generalize across a range of degraded conditions.

\begin{figure}[H]
\centering
\includegraphics[scale=0.5]{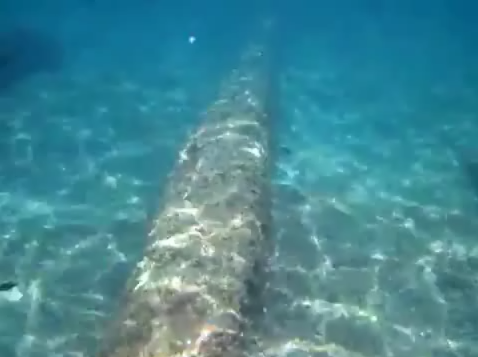}
\caption{Frame of one video of the first dataset.}
\label{fig:first_dataset}
\end{figure}

An example frame from one of these low-quality videos is shown in Figure~\ref{fig:first_dataset}. This figure illustrates the visual challenges faced in the dataset, including blurriness, uneven lighting, and low contrast.

\subsubsection{Second dataset: Underwater Caves Sonar and Vision Dataset}
The second dataset used for the experiments is the \textit{Underwater Caves Sonar and Vision Dataset} from the Underwater Vision and Robotics Lab at the University of Girona \cite{doi:10.1177/0278364917732838}. This dataset was collected with an autonomous underwater vehicle in the unstructured environment of the "Cuevas de Cala Viuda". The dataset was gathered in July 2013 and features challenging conditions for mapping and exploration tasks due to the confined, complex cave structures.

\begin{figure}[!ht]
\centering
\includegraphics[scale=0.5]{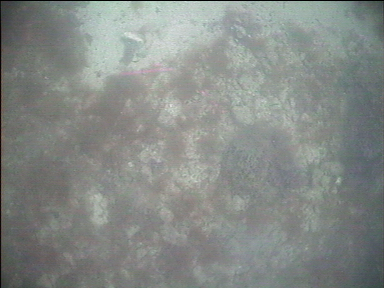}
\caption{Example of the underwater cave environment used for the Girona dataset.}
\label{fig:caves_dataset}
\end{figure}

An example frame from this dataset is shown in Figure \ref{fig:caves_dataset}, which highlights the unique conditions in underwater cave environments, including darkness, particulate matter, and intricate textures of the cave walls.

The dataset includes images captured from various sensors mounted on the AUV, including a down-looking camera used for ground truth validation. This dataset was chosen due to its application in a real-world scenario with a practical purpose, providing insights into the performance of the enhancement networks in authentic underwater environments. By utilizing data collected from actual field operations, we are able to evaluate the networks' effectiveness and potential for deployment in real-world use cases, ensuring that the results reflect the complexities and challenges of real underwater exploration.

\subsection{SuperGlue for Keypoint Matching}
SuperGlue is a crucial component in this study for matching keypoints between enhanced underwater images. It utilizes a graph neural network (GNN) architecture with self-attention and cross-attention mechanisms to improve the reliability of feature matching. This method builds upon the extracted keypoints and descriptors from the SuperPoint model, enhancing the feature-matching process by learning more discriminative matching descriptors through a neural network-based approach.

Before matching keypoints with SuperGlue, we use the SuperPoint model for feature extraction. SuperPoint is a pre-trained convolutional neural network designed for detecting keypoints and generating their corresponding descriptors. In the context of underwater images, SuperPoint identifies salient features such as corners and edges, while generating unique descriptors that represent the texture and appearance around each keypoint. The process workflow looks as follows:
\begin{enumerate}
    \item \textbf{Image enhancement}. Underwater images are first processed by one of the GAN-based enhancement networks (e.g., Funie-GAN, FSpiral-GAN).
    \item \textbf{Keypoint detection}. SuperPoint is applied to detect keypoints in the enhanced images.
    \item \textbf{Descriptor generation}. SuperPoint generates a descriptor for each detected keypoint, encapsulating information about its surrounding environment (texture, contrast, etc.).
    \item \textbf{Feature matching with SuperGlue}. The descriptors from SuperPoint are fed into SuperGlue, which matches keypoints between image pairs using its attention-based mechanism.
\end{enumerate}

\subsubsection{SuperGlue in Underwater Image Matching}
Underwater images often suffer from issues such as lighting distortions, color degradation, and blur, which can affect the consistency of keypoints across images. Traditional keypoint matching methods like ORB or SIFT struggle in these conditions due to the dynamic nature of underwater environments. SuperGlue, however, significantly improves the robustness of keypoint matching by leveraging its graph-based architecture and attention mechanisms.

\begin{figure}[h]
\centering
\includegraphics[scale=0.3]{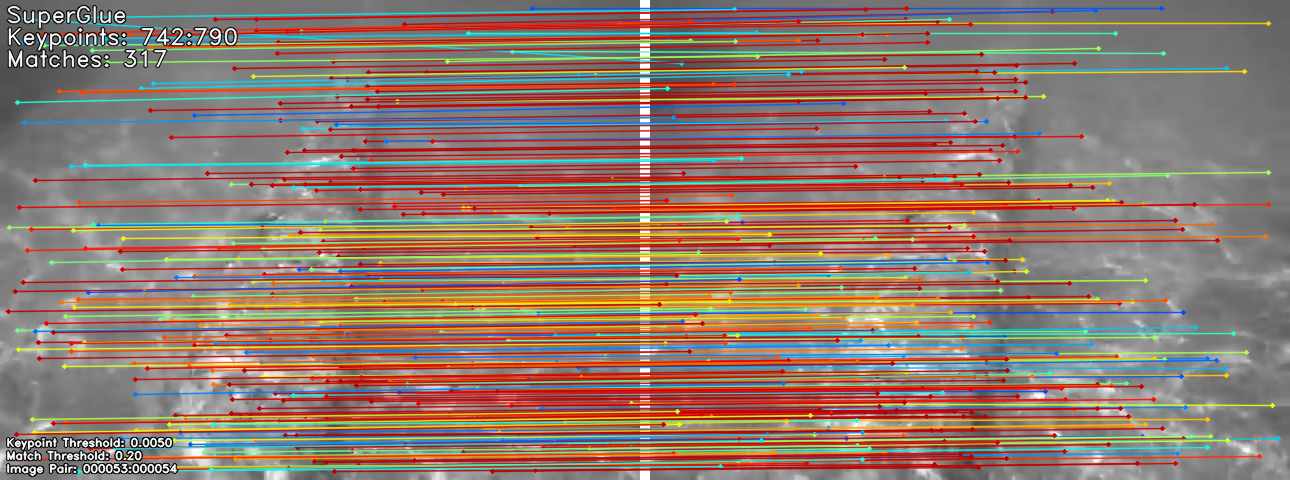}
\caption{SuperGlue feature matching between two enhanced frames. Color indicates confidence level, from low (blue) to high (red).}
\label{fig:SuperGlue_matching}
\end{figure}

For instance, in one of the experiments conducted with FSpiral-GAN-enhanced images, SuperGlue detected fewer keypoints compared to traditional methods, but the matches it identified were more accurate and reliable. An example of this process is shown in Figure~\ref{fig:SuperGlue_matching}. In this case, SuperGlue matched keypoints between two frames of a video, highlighting the improved reliability even in challenging conditions with caustics and turbidity.

In this particular example, the confidence level for matched keypoints exceeds 20\%, as indicated by the color gradient (from blue for lower confidence to red for higher confidence). While fewer keypoints were detected compared to traditional methods, SuperGlue achieved a matching accuracy of 42.7\% across 742 potential matches, identifying 317 correct matches.

\subsection{Evaluation Metrics}
\label{subsec:metrics}

As mentioned above, we employ several metrics to evaluate the performance of different keypoint detection and matching methods in underwater imagery. The main metrics employed are \textbf{AUC}, \textbf{Accuracy}, \textbf{Repeatability}, and \textbf{Match Score}. These metrics allow a comprehensive assessment of the degree to which each image enhancement method improves the accuracy and reliability of feature matching in degraded underwater images.

\begin{itemize}
    \item \textbf{AUC}. It represents the area under the Precision-Recall curve and is used to measure the performance of keypoint detection and matching at various thresholds. In our tests, we compute the AUC at three different thresholds: AUC@3, AUC@5, and AUC@10. These thresholds refer to the level of tolerance for errors in matching, where AUC@3 represents stricter matching requirements, and AUC@10 allows more flexibility. Higher AUC values indicate better overall performance in detecting and matching keypoints across a range of threshold settings.

    \item \textbf{Precision}. It measures the ratio of correctly matched keypoints to the total number of keypoints matched. It evaluates the accuracy of the feature matching process, indicating the percentage of correct matches out of all proposed matches. In underwater imagery, where distortions can lead to incorrect matches, a high precision score is crucial for reliable visual understanding and mapping. Precision is particularly important in applications like underwater navigation and structure-from-motion , where false matches can lead to significant errors.

    \item \textbf{Repeatability}. Repeatability, or REP, quantifies the ability of a method to consistently detect the same keypoints across multiple images or video frames under varying conditions. High repeatability indicates that a keypoint detector can reliably find the same features even when images undergo transformations such as changes in lighting, scale, or viewpoint. This metric is especially important in underwater environments, where such variations are common due to water conditions, movement of objects, and camera perspective.

    \item \textbf{Matching Score}. The Matching Score measures the overall quality of feature matches between two images. It considers both the number of correct matches and their geometric consistency, providing a holistic view of the matching performance. A higher MScore indicates that not only are the correct matches being identified, but they are also spatially consistent, which is essential for tasks such as 3D reconstruction and SLAM. This metric reflects the robustness of the matching process in dealing with underwater image challenges.
\end{itemize}

\subsection{Analysis of Keypoint Matching Performance}
\label{subsec:Test_first_dataset}

At first, we tested the networks on the first dataset. In the first video, which features challenging caustics, FSpiral-GAN demonstrated good overall performance, particularly in precision and repeatability. Although ORB achieved the highest AUC scores, it struggled in terms of repeatability. SIFT offered a moderate balance between AUC and repeatability, but its performance was lower than the deep learning-based models.

The results of the keypoint matching inhing in this video are presented in Table \ref{table:video1}. The visual challenges in this video caused traditional methods such as ORB and SIFT to perform poorly in terms of repeatability, while FSpiral-GAN demonstrated robustness.

\begin{table}[!ht]
\centering
\caption{Keypoint Matching Results for the First Video}
\label{table:video1}
\renewcommand{\arraystretch}{1.2} 
\small 
\begin{tabularx}{\textwidth}{|X|X|X|X|X|X|}
\hline
\textbf{Method} & \textbf{AUC@3} & \textbf{AUC@5} & \textbf{AUC@10} & \textbf{REP} & \textbf{n} \\ \hline
FSpiral-GAN & 44.60 & 44.97 & 46.78 & \textbf{48.81} & \textbf{1033.894} \\ \hline
SIFT & 62.58 & 63.44 & 65.48 & 23.44 & 907.610 \\ \hline
ORB & \textbf{77.39} & \textbf{78.46} & \textbf{80.88} & 7.43 & 499.586 \\ \hline
\end{tabularx}
\end{table}

In the second video, which lacked prominent caustics, FSpiral-GAN continued to show strong results. However, SIFT was second best in terms of repeatability (\textit{REP}), achieving a higher number of repeatable keypoints compared to the other methods.

\begin{table}[!ht]
\centering
\caption{Keypoint Matching Results for Video Number Two}
\label{table:video2}
\renewcommand{\arraystretch}{1.2} 
\small 
\begin{tabularx}{\textwidth}{|X|X|X|X|X|X|}
\hline
\textbf{Method} & \textbf{AUC@3} & \textbf{AUC@5} & \textbf{AUC@10} & \textbf{REP} & \textbf{n} \\ \hline
FSpiral-GAN & 46.72 & 48.28 & 51.55 & \textbf{47.17} & \textbf{796.129} \\ \hline
SIFT & 58.97 & 59.85 & 62.43 & 29.96 & \textbf{1604.75} \\ \hline
ORB & \textbf{70.53} & \textbf{71.62} & \textbf{73.96} & 9.73 & 499.854 \\ \hline
\end{tabularx}
\end{table}

In the third video, FSpiral-GAN again demonstrated superior repeatability, highlighting its strength in matching keypoints across different frames in an underwater environment with fewer caustics. ORB achieved the highest AUC scores, but its repeatability was significantly lower.

\begin{table}[!ht]
\centering
\caption{Keypoint Matching Results for Video Number Three}
\label{table:video3}
\renewcommand{\arraystretch}{1.2} 
\small 
\begin{tabularx}{\textwidth}{|X|X|X|X|X|X|}
\hline
\textbf{Method} & \textbf{AUC@3} & \textbf{AUC@5} & \textbf{AUC@10} & \textbf{REP} & \textbf{n} \\ \hline
FSpiral-GAN & 37.97 & 38.55 & 39.98 & \textbf{45.14} & \textbf{1362.355} \\ \hline
SIFT & 45.22 & 46.15 & 48.10 & 19.67 & 820.151 \\ \hline
ORB & \textbf{55.39} & \textbf{56.26} & \textbf{58.58} & 6.23 & 384.993 \\ \hline
\end{tabularx}
\end{table}

In the fourth video, featuring a ray on the seabed, FSpiral-GAN achieved the best overall repeatability, while ORB delivered higher AUC scores but struggled with consistency in repeatability.

\begin{table}[!ht]
\centering
\caption{Keypoint Matching Results for Video Number Four}
\label{table:video4}
\renewcommand{\arraystretch}{1.2} 
\small 
\begin{tabularx}{\textwidth}{|X|X|X|X|X|X|}
\hline
\textbf{Method} & \textbf{AUC@3} & \textbf{AUC@5} & \textbf{AUC@10} & \textbf{REP} & \textbf{n} \\ \hline
FSpiral-GAN & 35.03 & 35.82 & 37.39 & \textbf{56.79} & \textbf{534.921} \\ \hline
SIFT & 43.42 & 44.18 & 46.12 & 32.17 & 620.357 \\ \hline
ORB & \textbf{50.92} & \textbf{51.79} & \textbf{53.94} & 11.56 & 292.317 \\ \hline
\end{tabularx}
\end{table}

In the fifth video, which featured a ray and large caustics, FSpiral-GAN showed high repeatability and precision, while ORB achieved better AUC scores but again exhibited lower repeatability. While these methods delivered high AUC scores, especially for ORB, they exhibited lower repeatability and precision, particularly in more complex underwater scenes. This suggests that although traditional methods can perform well in certain scenarios, modern deep learning techniques like FSpiral-GAN and SuperPoint offer more robust and consistent results in environments with significant visual challenges, such as those present in underwater imagery.

\begin{table}[!ht]
\centering
\caption{Keypoint Matching Results for Video Number Five}
\label{table:video5}
\renewcommand{\arraystretch}{1.2} 
\small 
\begin{tabularx}{\textwidth}{|X|X|X|X|X|X|}
\hline
\textbf{Method} & \textbf{AUC@3} & \textbf{AUC@5} & \textbf{AUC@10} & \textbf{REP} & \textbf{n} \\ \hline
FSpiral-GAN & 38.41 & 38.93 & 40.78 & \textbf{60.02} & 190.727 \\ \hline
SIFT & 45.92 & 46.81 & 49.23 & 41.67 & \textbf{902.510} \\ \hline
ORB & \textbf{55.08} & \textbf{56.15} & \textbf{58.48} & 20.43 & 410.953 \\ \hline
\end{tabularx}
\end{table}

\subsubsection{Uncertainty Test with FSpiral-GAN}
GAN networks generate new videos from the original, which affects the input to SuperGlue. A global sensitivity analysis was conducted on the input parameters, specifically the video, to examine the impact of this variability. This variability is evident in the experimentation with GAN networks, where preprocessing significantly influences SuperGlue’s results. It becomes necessary to generate multiple videos with the same GAN network to achieve the optimal result.

This behavior leads to noticeable differences in the AUC, REP, or MScore values between videos generated by the same GAN network. To further explore this variability, we conducted an uncertainty test by generating 50 videos from the original first video using FSpiral-GAN. Each generated video was evaluated for key metrics, including AUC, precision, repeatability, and matching score. The table below presents the mean results across the 50 generated videos and compares them to the initial test result with a single video. The data suggest that the overall performance improves when considering the average of multiple generated videos, as seen in Table \ref{tab:uncertainty_fspiral_gan}.
\begin{table}[ht]
\centering
\caption{Uncertainty Test Results for FSpiral-GAN without Preprocessing}
\label{tab:uncertainty_fspiral_gan}
\begin{tabular}{|c|c|c|c|c|c|c|}
\hline
\textbf{AUC@3} & \textbf{AUC@5} & \textbf{AUC@10} & \textbf{Precision} & \textbf{REP} & \textbf{MScore} & \textbf{n} \\ \hline
45.29 & 45.88 & 47.52 & 99.94 & 48.86 & 48.80 & 1034.002 \\ \hline
\end{tabular}
\end{table}

\begin{figure}[h]
\centering
\includegraphics[scale=0.45]{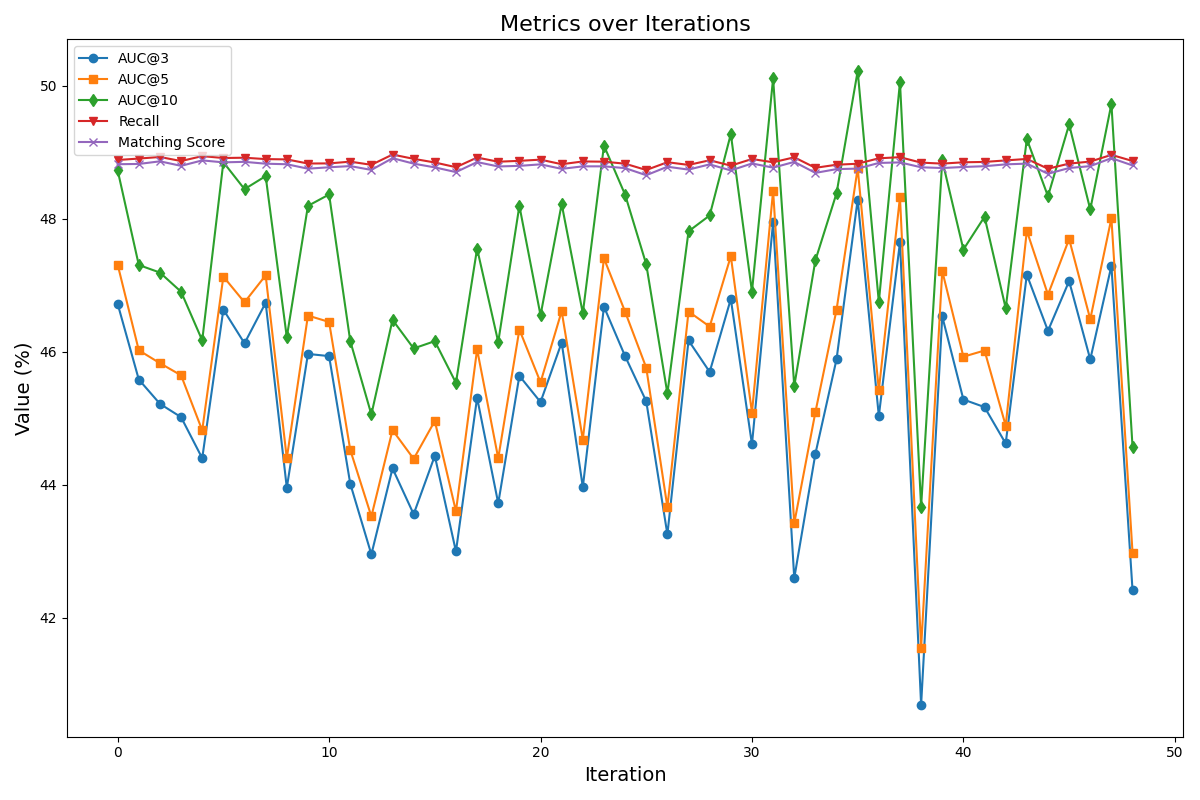}
\caption{Results obtained on the uncertainity test for FSpiral-GAN without preprocesing}
\label{fig:graph_fspiral_WP}
\end{figure}

The AUC3 value, as shown in Figure \ref{fig:graph_fspiral_WP}, has a maximum of 48.29, a minimum of 40.68, and a mean of 45.29, with an error range of +2.988\% to -4.60\%. Similarly, the AUC5 and AUC10 metrics present variability, but their average values are higher than those obtained from a single generated video.

This variability indicates that while FSpiral-GAN produces visually similar videos, they exhibit distinct characteristics when analyzed with SuperGlue. The sensitivity of AUC values to different video instances suggests that GAN-generated videos introduce variability in keypoint detection.

In terms of precision, the results show a narrow error margin, with a maximum precision of 99.95\% and a minimum of 99.92\%, indicating that FSpiral-GAN maintains a high degree of accuracy in keypoint matching. The repeatability and MScore metrics, while exhibiting slight variability, remain consistently high across the 50 videos, demonstrating that FSpiral-GAN is effective in identifying repeatable features in underwater imagery. 

To further investigate the impact of postprocessing techniques on keypoint detection, a series of image enhancement techniques were applied to the generated videos. These techniques included contrast enhancement, color correction, noise filtering, and superresolution. The goal was to recover lost image details and improve feature clarity, which may have been compromised during the resizing required by FSpiral-GAN. We will call the networks that will get this improvement FSpiral-GAN (Imp)  and Funie-GAN (Imp).

\begin{table}[ht]
\centering
\caption{Uncertainty Test Results for FSpiral-GAN with Postprocessing}
\label{tab:uncertainty_fspiral_gan_postprocessing}
\begin{tabular}{|c|c|c|c|c|c|}
\hline
\textbf{AUC@3} & \textbf{AUC@5} & \textbf{AUC@10} & \textbf{Precision} & \textbf{REP} & \textbf{MScore} \\ \hline
47.00 & 47.68 & 49.52 & 99.96 & 49.74 & 49.61  \\ \hline
\end{tabular}
\end{table}

\begin{figure}[h]
\centering
\includegraphics[scale=0.45]{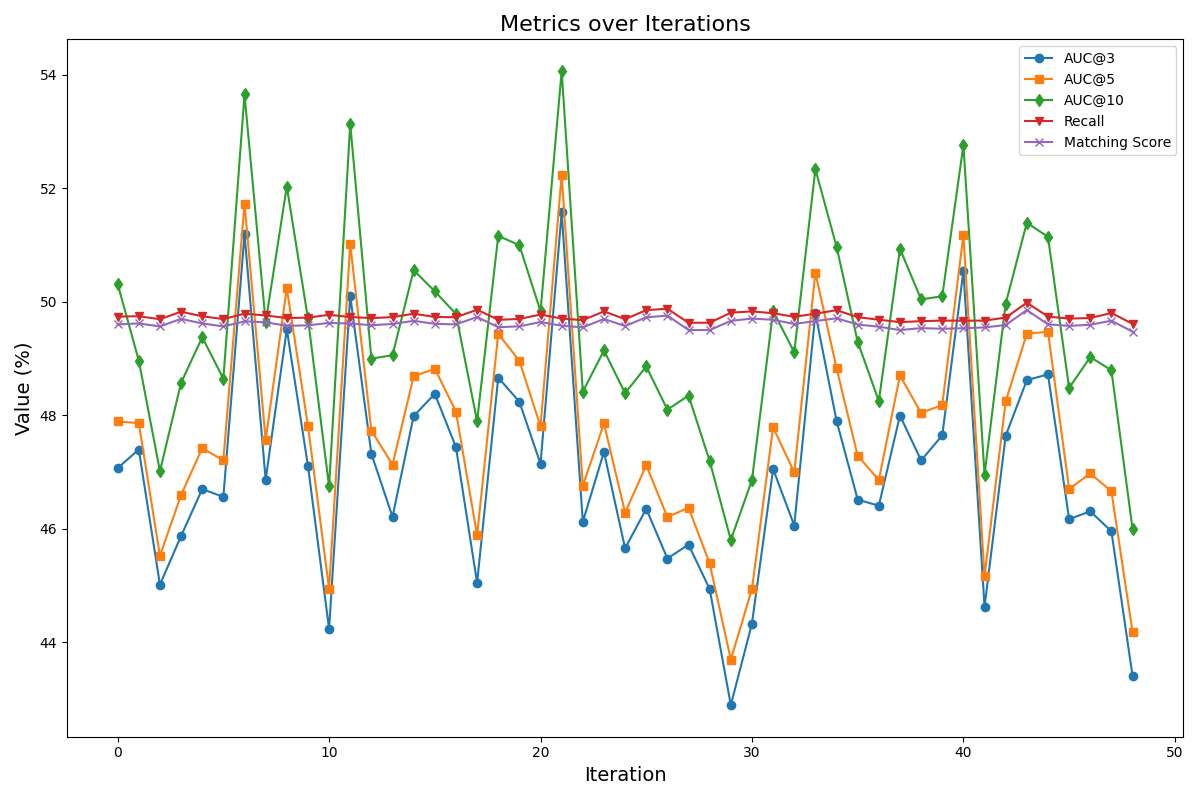}
\caption{Results obtained on the uncertainity test for FSpiral-GAN with preprocesing}
\label{fig:graph_fspiral_P}
\end{figure}

The AUC3, AUC5, and AUC10 metrics, as shown in \ref{fig:graph_fspiral_P}, after postprocessing show that the error margins decrease slightly compared to the non-postprocessed results. Additionally, REP, precision and MScore exhibit more consistent results indicating that the postprocessing steps successfully enhance the repeatability and reliability of keypoint matching across the video frames. This suggests that postprocessing can mitigate some of the variability introduced by GAN networks, enhancing the overall quality of keypoint detection.

The reason we move away from comparisons with traditional methods like SIFT and ORB is that the results obtained with FSpiral-GAN already highlight their limitations. In underwater environments, where image distortions such as caustics and blurring are prevalent, SIFT and ORB struggle to detect reliable keypoints and produce consistent matches, as we said on the SupeGlue section. In contrast, deep learning-based methods, particularly when combined with SuperPoint and SuperGlue, offer far more robust results. The initial experiments with FSpiral-GAN show that traditional methods not only underperform in precision and repeatability but also fail to handle the complexities introduced by enhanced underwater images. Therefore, continuing with SIFT or ORB would not provide meaningful improvements, which is why the focus shifts entirely towards advanced deep learning models for the remainder of the experimentation.

\subsubsection{Uncertainty Analysis on other networks}

To determine whether the variability observed in FSpiral-GAN extends to other GAN networks, we conducted the same uncertainty test on Funie-GAN. Interestingly, Funie-GAN produces consistent results across all 50 generated videos. The lack of variability suggests that Funie-GAN does not introduce the same level of randomness into the generated videos, resulting in more uniform keypoint detection outcomes. The following table \ref{tab:uncertainty_all} shows the final result of all the AIs on this test.

\begin{table}[ht]
\centering
\caption{Results with the uncertainty analysis test}
\renewcommand{\arraystretch}{1.2} 
\small 
\begin{tabularx}{\textwidth}{|X|X|X|X|X|X|X|}
\hline
\textbf{Method} & \textbf{AUC@3} & \textbf{AUC@5} & \textbf{AUC@10} & \textbf{Precision} & \textbf{REP} & \textbf{MScore}  \\ \hline
FSpiral-GAN & 45.29 & 45.88 & 47.52 & 99.94 & 48.86 & 48.80 \\ \hline
FSpiral-GAN (Imp) & 47.00 & 47.68 & 49.52 & \textbf{99.96} & 49.74 & 49.61  \\ \hline
Funie-GAN & 42.75 & 43.48 & 45.31 & 99.75 & 46.64 & 46.66 \\ \hline
Funie-GAN (Imp) & 42.48 & 43.24 & 44.67 & 99.91 & 48.69 & 48.34 \\ \hline
UWCNN & \textbf{48.49} & \textbf{49.27} & \textbf{51.24} & 99.54 & 46.21 & 46.14 \\ \hline
U-NET & 48.10 & 48.47 & 49.80 & 99.59 & 49.85 & 49.75 \\ \hline
FA-NET & 45.76 & 46.24 & 47.75 & 99.82 & \textbf{52.43} & \textbf{52.49} \\ \hline
\end{tabularx}
\label{tab:uncertainty_all}
\end{table}

\begin{figure}[h]
    \centering
    \begin{subfigure}[b]{0.45\textwidth}
        \centering
        \includegraphics[width=\textwidth]{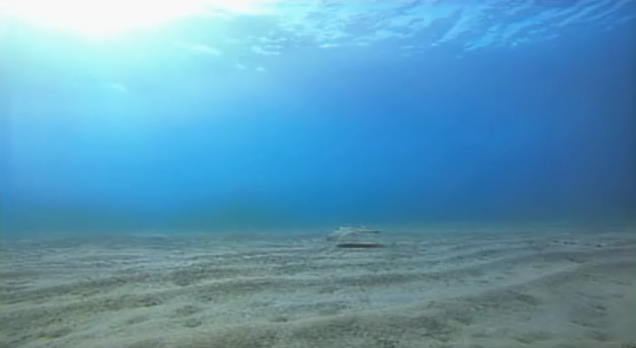}
        \caption{Funie-GAN processed frame.}
        \label{fig:funiegan}
    \end{subfigure}
    \hfill
    \begin{subfigure}[b]{0.45\textwidth}
        \centering
        \includegraphics[width=\textwidth]{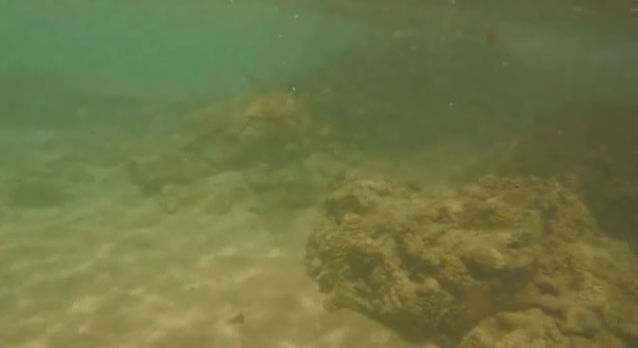}
        \caption{FSpiral-GAN processed frame.}
        \label{fig:fspiralgan}
    \end{subfigure}
    
    \caption{Comparison of processed frames from two GAN models: (a) Funie-GAN and (b) FSpiral-GAN.}
    \label{fig:gan_comparison}
\end{figure}

Funie-GAN shows steady performance in precision, REP, and MScore. These consistent results across multiple videos indicate that Funie-GAN is a reliable choice when consistent outputs are required. However, the overall performance remains slightly lower than that of FSpiral-GAN, particularly in the AUC and REP metrics. When postprocessing is applied to Funie-GAN’s output (Table \ref{tab:uncertainty_all}), there is a slight improvement in the results, but they still fall short of the performance achieved by FSpiral-GAN with postprocessing. We can see some images of procesed images of FSpiral and Funie-GAN in Figure \ref{fig:gan_comparison}.

These results suggest that, while Funie-GAN is less sensitive to variability, its overall performance is not as robust as that of FSpiral-GAN. Nonetheless, it remains a solid choice for applications where consistent outputs are required.

\begin{figure}[h]
    \centering
    \begin{subfigure}[b]{0.32\textwidth}
        \centering
        \includegraphics[width=\textwidth, height = 3cm]{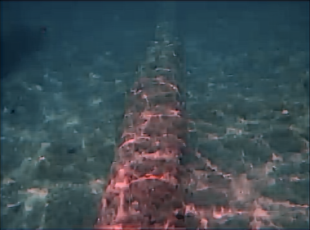}
        \caption{UWCNN.}
        \label{fig:uwcnn}
    \end{subfigure}
    \hfill
    \begin{subfigure}[b]{0.32\textwidth}
        \centering
        \includegraphics[width=\textwidth, height = 3cm]{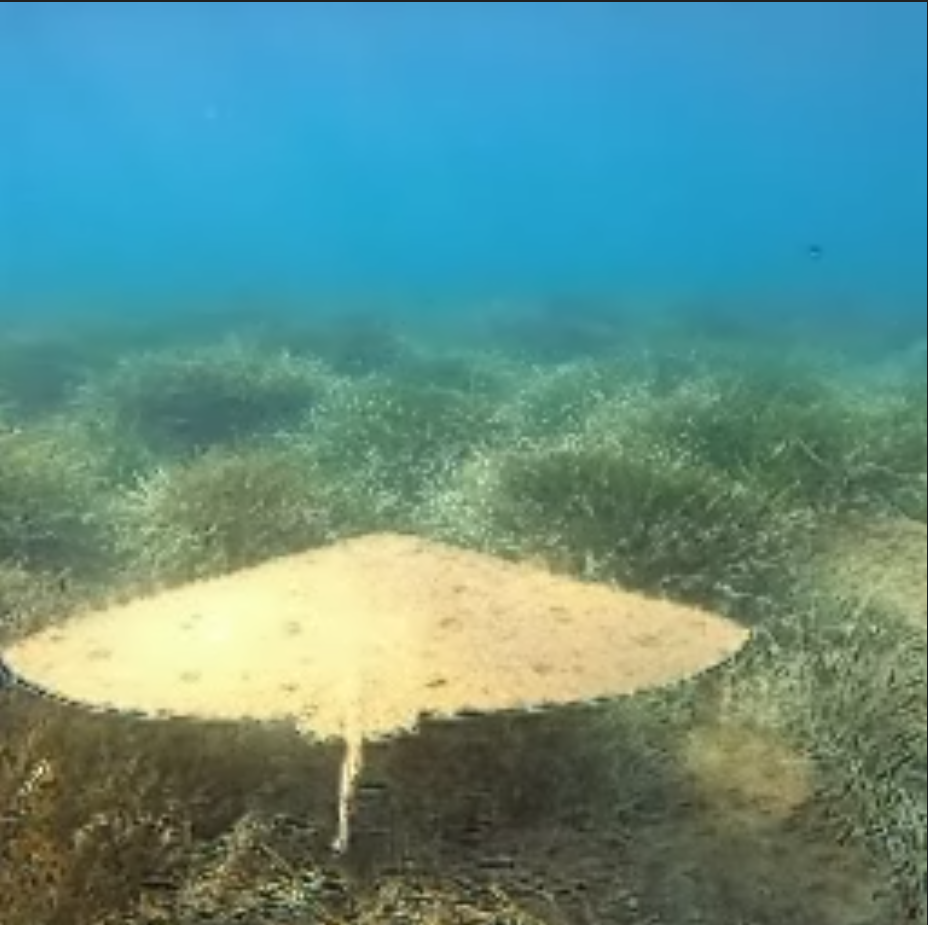}
        \caption{U-Net.}
        \label{fig:unet}
    \end{subfigure}
    \hfill
    \begin{subfigure}[b]{0.32\textwidth}
        \centering
        \includegraphics[width=\textwidth, height = 3cm]{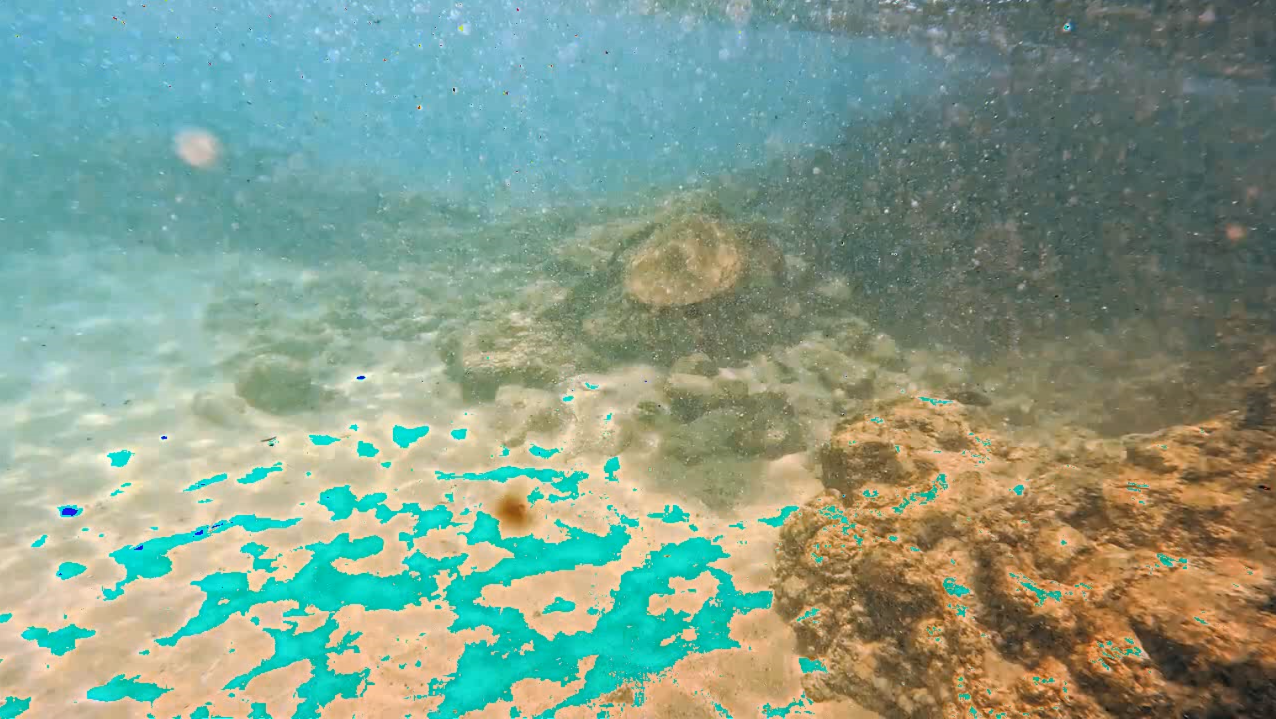}
        \caption{Five A+.}
        \label{fig:fivea}
    \end{subfigure}
    
    \caption{Example of images using during the test: (a) UWCNN, (b) U-Net, and (c) Five A+.}
    \label{fig:U-net_UWCNN_Five A+}
\end{figure}

The uncertainty analysis for UWCNN and U-Net reveals that these networks produce consistent results, similar to Funie-GAN. The metrics do not show significant variability across the 50 generated videos, as seen in Table \ref{tab:uncertainty_all}. Both networks demonstrate strong performance in terms of AUC and precision, with U-Net slightly outperforming UWCNN in repeatability and MScore. We can see some images of processed images of UWCNN, U-Net and Five A+ in Figure \ref{fig:U-net_UWCNN_Five A+}.

The results indicate that U-Net offers better overall performance compared to UWCNN, especially in repeatability and MScore. However, both networks perform well across different thresholds, making them reliable options for feature matching in underwater environments. Their consistent outputs make them suitable for tasks requiring high reliability without the variability seen in GAN-based networks.

\subsubsection{Keypoint Matching Results for the second dataset}
\label{subsec:Test_second_dataset}
In order to see the performance on a real world enviorement with high quality images we searched the \textit{Underwater Caves Sonar and Vision Dataset}. This dataset was used to obtain the results from applying the different image enhancement networks seen before to it. The main goal is to assess the performance of these networks using feature matching tests, where we evaluate their ability to enhance image quality and facilitate the identification of common features. As well as the last section the evaluation metrics include AUC, precision, repeatability, and matching score. 

\begin{table}[!ht]
\centering
\caption{Keypoint Matching Results for the Underwater Caves Dataset}
\renewcommand{\arraystretch}{1.2} 
\small 
\begin{tabularx}{\textwidth}{|X|X|X|X|X|X|X|}
\hline
Network & AUC@3 & AUC@5 & AUC@10 & Precision & REP & MScore \\ \hline
Original dataset & 66.38  &  67.30  & 69.32 &  98.07  & 36.09  & 35.38 \\ \hline
FSpiral-GAN & 60.18 & 61.27 & 63.94 & 99.56 & 40.81 & 40.37 \\ \hline
FSpiral-GAN (Imp) & 68.02  & 68.98  & 70.99 & 97.63 & 39.87 & 38.29 \\ \hline
Funie-GAN & 42.75 & 43.48 & 45.31 & \textbf{99.75} & \textbf{46.64} & \textbf{46.66} \\ \hline
Funie-GAN (Imp) & \textbf{69.70} &  \textbf{70.84} & \textbf{73.15}  & 97.24 & 38.14 & 36.31 \\ \hline
UWCNN & 58.14  & 59.26 & 61.91 & 98.72 & 37.40  & 37.40 \\ \hline
U-Net & 60.25  & 61.50 & 64.58 & 98.45 & 41.94  & 40.91 \\ \hline
FA-Net & 53.47 & 54.47 & 57.43 & 99.68 & 42.88 & 41.90 \\ \hline
\end{tabularx}
\label{tab:underwater_cave_results}
\end{table}

The results obtained from the underwater cave dataset can be found in Table \ref{tab:underwater_cave_results}. The original dataset achieves \textit{AUC} values between 66.38 and 69.32, but has a lower precision of 98.07 and a relatively low \textit{MScore} of 35.38. This indicates that, without improvement, the dataset has difficulty in systematically identifying high quality data. This indicates that, without improvements, the dataset has difficulty in systematically identifying high quality key points in the challenging underwater environment. It therefore serves as a benchmark for comparison.

By applying \textit{FSpiral-GAN}, the network shows an improved accuracy of 99.56 and a significant increase in repeatability (\textit{REP} of 40.81), outperforming the original data set. These results suggest that \textit{FSpiral-GAN} is effective in improving the accuracy of feature detection in underwater images. However, the \textit{AUC} values are slightly lower than the baseline, indicating that the network may have difficulty distinguishing features between different thresholds. Post-processing of the \textit{FSpiral-GAN} images leads to a marked improvement in \textit{AUC}, with values reaching as high as 70.99, although this comes at the expense of a small reduction in accuracy (97.63). This trade-off suggests that, although post-processing improves overall detection, some details may be lost.

\textit{Funie-GAN} exhibits a different performance profile, with lower \textit{AUC} values ranging from 42.75 to 45.31, indicating weaker performance in feature detection across multiple thresholds. However, it outperforms in accuracy, reaching 99.75, and a strong repeatability score (\textit{REP} of 46.64), making it very effective when accuracy is the primary concern. After post-processing, \textit{Funie-GAN} improves on \textit{AUC}, reaching the highest values among the tested networks (up to 73.15), although the accuracy decreases slightly to 97.24. Overall, \textit{Funie-GAN} is shown to be very effective in underwater image enhancement, especially when post-processing techniques are applied.

Both \textit{UWCNN} and \textit{U-Net} offer balanced performance. \textit{UWCNN} achieves slightly lower accuracy at 98.72, but its repeatability and \textit{MScore} suggest consistency in feature detection. In contrast, \textit{U-Net} performs well in all metrics, with \textit{AUC} values between 60.25 and 64.58, an accuracy of 98.45, and a \textit{REP} of 41.94. These results highlight the reliability of \textit{U-Net} in detecting key points even in degraded underwater images.

\textit{FA-Net} shows intermediate results, with \textit{AUC} values ranging from 53.47 to 57.43. However, it stands out in accuracy (\textit{AUC}). Furthermore, it improves in precision (99.68) and repeatability (\textit{REP} of 42.88). Although its \textit{AUC} scores are not as high as some of the other networks, \textit{FA-Net} is consistent and reliable feature detection, especially in challenging environments where high accuracy and repeatability are paramount.

\begin{table}[H]
\centering
\caption{Keypoint Matching Results for Experiment 1 (Low-Quality Videos)}
\renewcommand{\arraystretch}{1.2} 
\small 
\label{tab:experiment1_results}
\begin{tabularx}{\textwidth}{|X|X|X|X|X|}
\hline
\textbf{Method} & \textbf{AUC@3} & \textbf{AUC@5} & \textbf{AUC@10} & \textbf{REP}  \\ \hline
FSpiral-GAN (Video 1) & 44.60 & 44.97 & 46.78 & \textbf{48.81}  \\ \hline
SIFT (Video 1) & 62.58 & 63.44 & 65.48 & 23.44  \\ \hline
ORB (Video 1) & \textbf{77.39} & \textbf{78.46} & \textbf{80.88} & 7.43  \\ \hline
FSpiral-GAN (Video 2) & 46.72 & 48.28 & 51.55 & \textbf{47.17}  \\ \hline
SIFT (Video 2) & 58.97 & 59.85 & 62.43 & 29.96  \\ \hline
ORB (Video 2) & \textbf{70.53} & \textbf{71.62} & \textbf{73.96} & 9.73  \\ \hline
FSpiral-GAN (Video 3) & 37.97 & 38.55 & 39.98 & \textbf{45.14}  \\ \hline
SIFT (Video 3) & 45.22 & 46.15 & 48.10 & 19.67  \\ \hline
ORB (Video 3) & \textbf{55.39} & \textbf{56.26} & \textbf{58.58} & 6.23  \\ \hline
FSpiral-GAN (Video 4) & 35.03 & 35.82 & 37.39 & \textbf{56.79} \\ \hline
SIFT (Video 4) & 43.42 & 44.18 & 46.12 & 32.17  \\ \hline
ORB (Video 4) & \textbf{50.92} & \textbf{51.79} & \textbf{53.94} & 11.56  \\ \hline
FSpiral-GAN (Video 5) & 38.41 & 38.93 & 40.78 & \textbf{60.02} \\ \hline
SIFT (Video 5) & 45.92 & 46.81 & 49.23 & 41.67  \\ \hline
ORB (Video 5) & \textbf{55.08} & \textbf{56.15} & \textbf{58.48} & 20.43  \\ \hline
\end{tabularx}
\end{table}
\FloatBarrier

\begin{table}[H]
\centering
\caption{Results with the uncertainty analysis test}
\renewcommand{\arraystretch}{1.2} 
\small 
\begin{tabularx}{\textwidth}{|X|X|X|X|X|X|X|}
\hline
\textbf{Method} & \textbf{AUC@3} & \textbf{AUC@5} & \textbf{AUC@10} & \textbf{Precision} & \textbf{REP} & \textbf{MScore}  \\ \hline
FSpiral-GAN & 45.29 & 45.88 & 47.52 & 99.94 & 48.86 & 48.80 \\ \hline
FSpiral-GAN (Imp) & 47.00 & 47.68 & 49.52 & \textbf{99.96} & 49.74 & 49.61  \\ \hline
Funie-GAN & 42.75 & 43.48 & 45.31 & 99.75 & 46.64 & 46.66 \\ \hline
Funie-GAN (Imp) & 42.48 & 43.24 & 44.67 & 99.91 & 48.69 & 48.34 \\ \hline
UWCNN & \textbf{48.49} & \textbf{49.27} & \textbf{51.24} & 99.54 & 46.21 & 46.14 \\ \hline
U-NET & 48.10 & 48.47 & 49.80 & 99.59 & 49.85 & 49.75 \\ \hline
FA-NET & 45.76 & 46.24 & 47.75 & 99.82 & \textbf{52.43} & \textbf{52.49} \\ \hline
\end{tabularx}
\label{tab:uncertainty_all_summary}
\end{table}
\FloatBarrier

\begin{table}[H]
\centering
\caption{Keypoint Matching Results for the Underwater Caves Dataset}
\renewcommand{\arraystretch}{1.2} 
\small 
\begin{tabularx}{\textwidth}{|X|X|X|X|X|X|X|}
\hline
Network & AUC@3 & AUC@5 & AUC@10 & Precision & REP & MScore \\ \hline
Original dataset & 66.38  &  67.30  & 69.32 &  98.07  & 36.09  & 35.38 \\ \hline
FSpiral-GAN & 60.18 & 61.27 & 63.94 & 99.56 & 40.81 & 40.37 \\ \hline
FSpiral-GAN (Imp) & 68.02  & 68.98  & 70.99 & 97.63 & 39.87 & 38.29 \\ \hline
Funie-GAN & 42.75 & 43.48 & 45.31 & \textbf{99.75} & \textbf{46.64} & \textbf{46.66} \\ \hline
Funie-GAN (Imp) & \textbf{69.70} &  \textbf{70.84} & \textbf{73.15}  & 97.24 & 38.14 & 36.31 \\ \hline
UWCNN & 58.14  & 59.26 & 61.91 & 98.72 & 37.40  & 37.40 \\ \hline
U-Net & 60.25  & 61.50 & 64.58 & 98.45 & 41.94  & 40.91 \\ \hline
FA-Net & 53.47 & 54.47 & 57.43 & 99.68 & 42.88 & 41.90 \\ \hline
\end{tabularx}
\label{tab:underwater_cave_results_summary}
\end{table}
\FloatBarrier

\section{Conclusions}
\label{sec:conclusions}
In conclusion, the experiments on both synthetic and real-world underwater datasets provide valuable insights into the performance of various deep learning models for image enhancement in challenging environments. The original dataset, serving as a baseline, highlights the inherent difficulties in keypoint detection in underwater imagery, with lower precision and matching scores due to degraded visual conditions.

Among the tested networks, \textit{FSpiral-GAN} demonstrates strong performance, especially in terms of precision and repeatability. Its ability to significantly improve keypoint matching, particularly after post-processing, makes it a reliable solution for underwater image enhancement, although the slight trade-off in precision due to post-processing needs careful consideration.

\textit{Funie-GAN}, while starting with lower \textit{AUC} scores, shines in precision and repeatability, especially in the raw form. When post-processed, it achieves the highest \textit{AUC} scores across all networks, indicating its robustness in improving underwater image clarity and feature detection. This makes \textit{Funie-GAN} particularly effective for use cases where precision and consistent results are prioritized.

Both \textit{UWCNN} and \textit{U-Net} exhibit balanced performance across the different metrics. \textit{U-Net}, in particular, shows high reliability with strong results in repeatability and matching score, making it a versatile choice for underwater feature detection. \textit{FA-Net}, while not leading in \textit{AUC}, stands out with exceptional precision and repeatability, making it a strong candidate for applications requiring reliable feature consistency.

Overall, the experiments demonstrate that while traditional methods like SIFT and ORB can achieve high \textit{AUC} scores, they fall short in terms of repeatability and precision, especially in more complex underwater scenarios. In contrast, deep learning-based approaches such as \textit{FSpiral-GAN}, \textit{Funie-GAN}, and \textit{U-Net} offer more robust and consistent results, particularly in challenging underwater environments. These networks hold significant potential for practical applications in underwater exploration and monitoring, where consistent and accurate feature detection is important.

\section*{Considerations}
During the preparation of this work the authors used ChatGPT and DeepL in order to improve its readability. After using this services, the authors reviewed and edited the content as needed and take full responsibility for the content of the publication.

\bibliographystyle{plainnat}
\bibliography{main.bib}

\end{document}